\documentclass[10pt,twocolumn,letterpaper]{article}

\usepackage{cvpr}
\usepackage{times}
\usepackage{epsfig}
\usepackage{graphicx}
\usepackage{amsmath}
\usepackage{amssymb}

\usepackage{pifont}% http://ctan.org/pkg/pifont

\usepackage{color}
\usepackage{multirow}
\usepackage{caption}
\usepackage{subcaption}
\usepackage{tabularx} 
\usepackage{algorithm}  
\usepackage{algorithmic} 

\usepackage{booktabs}

\usepackage{amsthm}

\theoremstyle{definition}

\theoremstyle{remark}

\usepackage{algorithm}  
\usepackage{algorithmic} 

\usepackage{flushend} 
\usepackage[pagebackref=true,breaklinks=true,letterpaper=true,colorlinks,bookmarks=false]{hyperref}

\cvprfinalcopy % *** Uncomment this line for the final submission

\def\cvprPaperID{216} % *** Enter the CVPR Paper ID here

\ifcvprfinal\fi
\begin{document}

%%%%%%%%% TITLE
\title{Object Relation Detection Based on One-shot Learning}

\author{\normalsize{Li~Zhou \quad Jian~Zhao  \quad Jianshu~Li \quad Li~Yuan \quad Jiashi~Feng}\\
	\small{National University of Singapore } \\
	{\small zhouli2025@gmail.com \quad \{zhaojian90, jianshu, e0204947\}@u.nus.edu \quad elefjia@nus.edu.sg}}

\maketitle
%\thispagestyle{empty}

%%%%%%%%% ABSTRACT

\begin{abstract}
Detecting the relations among objects, such as ``cat on sofa'' and ``person ride horse'', is a crucial task in image understanding, and beneficial to bridging the semantic gap between images and natural language. Despite the remarkable progress of deep learning in detection and recognition of individual objects, it is still a challenging task to localize and recognize the relations between objects due to the complex combinatorial nature of various kinds of object relations. Inspired by the recent advances in one-shot learning, we propose a simple yet effective Semantics Induced Learner (SIL) model for solving this challenging task. Learning in one-shot manner can enable a detection model to adapt to  a huge number of object relations with diverse appearance effectively and robustly. In addition, the SIL combines bottom-up and top-down attention mechanisms, therefore enabling attention at the level of vision and semantics favorably. Within our proposed model, the bottom-up mechanism, which is based on Faster R-CNN, proposes objects regions, and the top-down mechanism selects and integrates visual features according to semantic information. Experiments demonstrate the effectiveness of our framework over other state-of-the-art methods on two large-scale data sets for object relation detection.
\end{abstract}
%\footnote{This is an abstract footnote}

%
% The code below should be generated by the tool at
% http://dl.acm.org/ccs.cfm
% Please copy and paste the code instead of the example below.
%

%\begin{CCSXML}
%<ccs2012>
% <concept>
%  <concept_id>10010520.10010553.10010562</%concept_id>
%  <concept_desc>Computer systems %organization~Embedded systems</concept_desc>
%  <concept_significance>500</concept_significance>
% </concept>
% <concept>
%  <concept_id>10010520.10010575.10010755</%concept_id>
%  <concept_desc>Computer systems %organization~Redundancy</concept_desc>
%  <concept_significance>300</concept_significance>
% </concept>
% <concept>
%  <concept_id>10010520.10010553.10010554</%concept_id>
%  <concept_desc>Computer systems %organization~Robotics</concept_desc>
%  <concept_significance>100</concept_significance>
% </concept>
% <concept>
%  <concept_id>10003033.10003083.10003095</%concept_id>
%  <concept_desc>Networks~Network reliability</%concept_desc>
%  <concept_significance>100</concept_significance>
% </concept>
%</ccs2012>
%\end{CCSXML}

%\ccsdesc[500]{Computer systems %organization~Embedded systems}
%\ccsdesc[300]{Computer systems %organization~Redundancy}
%\ccsdesc{Computer systems organization~Robotics}
%\ccsdesc[100]{Networks~Network reliability}
%\ccsdesc[100]{Networks~Network reliability}

\maketitle

\section{Introduction}

% So relationships are usually defined as tuples, which comprise of three parts: a subject, a predicate and an object. 

\begin{figure}

\includegraphics[width=8cm,height=8cm]{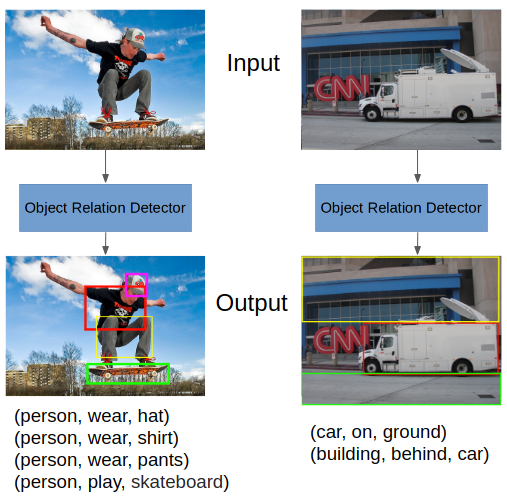}
\caption{Here are two examples from the Visual Relationship Dataset~\cite{lu2016visual}. Object relations widely exist  in real-world images, and their detection potentially benefits many applications. We develop a framework that can effectively localize and recognize such relations in a given image. As shown in the figure, the detected relations contain three parts: \emph{subject}, \emph{predicate} and \emph{object}.}
\label{figure1}
\end{figure}

Object relation is   abstract representation of the visually
observable interactions between a pair of a subject and an object, such as ``person play piano''. Detecting object relations in images is one crucial task in image understanding. Each object relation involves a pair of localized objects (subject and object) which are connected via a predicate. A predicate can be action (\emph{e.g.} ``play''), a spatial preposition (\emph{e.g.} ``on'') or some comparative expression (\emph{e.g.} ``taller than''). While objects are the basic constituent elements of an image, it is often the  relations between objects that provide the holistic interpretation of a scene. See \ref{figure1} for an illustration. Extracting such visual information would benefit many related multimedia  applications such as image captioning \cite{anderson2016spice, aditya2015images}, image retrieval \cite{johnson2015image} and visual question answering \cite{wu2017visual}.

\begin{figure*}
\includegraphics[width=18cm,height=4cm]{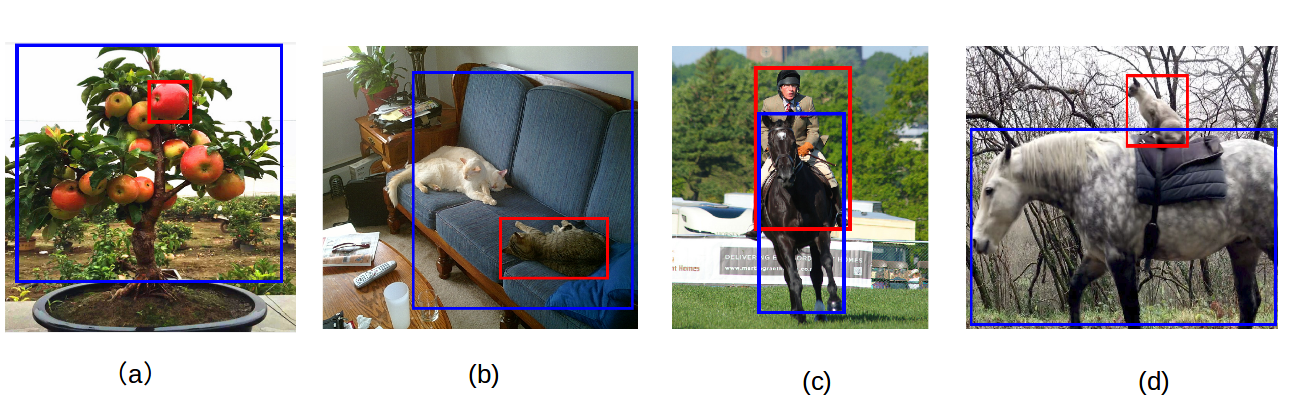}
\caption{(a) and (b) are examples of different semantics of the same predicates with different subjects and objects. (c) and (d) are examples of the same semantics of different predicates with different subjects and objects.}
\label{figure2}
\end{figure*}

% A natural solution to $long$-$tail$ problem is to learn two models for objects and predicates, respectively. In this case, complexity would be reduced to $O(N +R)$ from $O(N^2R)$. However, the learning becomes even more difficult due to the drastic appearance change of predicates. That is to say, the relationships which have the same predicate but different subjects or objects are treated as the same relationship type, resulting in ignorance of the important semantic dependencies between the subjects or objects and the predicates. For example, this solution regards the relationships ($apple$, $on$, $tree$) and ($cat$, $on$, $sofa$)as the same relationship type, because they both have the same predicate ``on". However, in human natural language, these two relationships delivery totally different semantic information. We term this challenge

Despite the exciting development in the efforts devoted to bridging natural language processing and computer vision \cite{antol2015vqa, bernardi2016automatic, ren2015faster}, it is still challenging and difficult to detect and understand  various object relations involved within multiple objects of an image. The difficulties can be attributed to the following three reasons. The first reason is the \textbf{highly complex combinatorial nature}. Given $N$ objects and $R$ predicates, a relation detection model has to examine $O(N^2 R)$ relations. These would lead to a huge number of potential relation types in real-world applications. For example, there exist more than 75K relation types in the Visual Genome dataset~\cite{krishna2017visual}. Meanwhile, each relation involves two parts, namely \emph{subject} and \emph{object}, resulting in a greater skew of rare relations especially when co-occurrence of some pairs of objects is infrequent in the dataset. Some types of relations contain very limited examples. We call this phenomenon the \textbf{long-tail} problem. The second reason is the \textbf{large intra-class divergence}. Relations that have the same predicate but different subjects or objects are essentially different. For example, (apple, on, tree) and (cat, on, sofa) are totally different. See \ref{figure2} (a) and (b) for  illustration. The third reason is the \textbf{semantic dependency} that the predicate in a relation is not only determined by semantic information but also the categories of the subject and the object. Taking \ref{figure2} (c) and (d) as examples, the relations between person and horse and between cat and horse are visually similar, but it is more common to say (cat, sit on, horse) instead of (cat, ride, horse) in  natural language. Similarly, it is also not often to say (person, sit on, horse).

%In this case, the same relationships would dilivery relatively similar semantics and the number of relationships is acceptable. However, such practice would unavoidably face an immense difficulty – it has to address the learning problem resulting from a great majority of imbalanced classes within dataset. For more details, Visual Genome dataset consists of more than 75K visual relationship types, but the number of examples for each kind of relationship varies from about several hundreds to more than ten thousands. In order to overcome the learning difficulty due to such a severely imbalanced distribution within dataset, how sophisticated need the classifier being designed?

In essence, object relation detection can be regarded as a classification task. Then the fundamental problem is how to formulate the relation triplet (subject, predicate, object) to define a reasonable classification task. Previous attempts~\cite{sadeghi2011recognition} used to treat the relation triplet (subject, predicate, object) as a whole to classify, but this strategy suffers from imbalanced relation examples and does not perform well for detecting rare relations. 
% - that is, learning separate models for objects and predicate - is another common solution. Object relations are compositions of objects and predicates, so the number of relationship types becomes equal to the number of predicate types because of the split. Even if the number of relationship types is reduced sharply - in the example of $N$ objects and $R$ predicates, the learning complexity is reduced to $O(N + R)$, quite a few predicates with totally different semantics in the original and complete relationship triplets $(subject,$ $predicate,$ $object)$ would be considered as the same class, just because those predicates are the same literally, e.g. $(cat,$ $on,$ $sofa)$ and $(apple,$ $on,$ $tree)$
Another strategy is to pick out the predicates from relation triplets by splitting (subject, predicate, object)  to two parts: predicate and subject/object. However, this would introduce a high diversity within each class, namely the \emph{intra-class divergence}, and bring extreme difficulty for model learning. The main reason why such a diversity arises is the ignorance of the important components of a relation triplet: subject and object, that is, the semantics delivered by the categories of objects. 

Most existing object relation detection works~\cite{zhang2017visual, lu2016visual} are only of the bottom-up variety. Taking as input the appearance representation of relation regions which are generally generated from object bounding boxes detected by a detector, these works, however, fail to put enough attention to the categories of the subject and object, and they ignore the important {semantics dependencies}. Without considering semantics dependencies, learning an object detection model would be challenged by two difficulties. The first difficulty is due to the intra-class divergence problem we discuss above. A relation may contain different semantics that would confuse the model. Second, the ignorance of semantics dependencies consequently results in the lack of attention of the model to visual feature regions at the high level. However, in the human visual system, one can concentrate his attention volitionally due to top-down signals guided by the current task (e.g., attempting to determine the relation between objects), and automatically due to bottom-up signals caused by salient or eye-catching stimuli.%~\cite{buschman2007top, corbetta2002control}.

According to the above observations, we propose to learn  models for objects and predicates respectively and develop a framework which combines both bottom-up and top-down attention mechanisms. We refer to the mechanism related to semantics dependencies as the \textbf{top-down mechanism} and the one purely related to visual representations as the \textbf{bottom-up mechanism}. The bottom-up mechanism generates a set of object proposals with category information, then visual features are extracted from these proposals by a Convolution Neural Network~\cite{simonyan2014very}. Practically, we implement bottom-up attention utilizing Faster R-CNN~\cite{ren2015faster}. The top-down mechanism exploits semantics dependencies (the categories of subjects and objects) to predict an attention distribution over the visual features from the bottom-up mechanism.

Besides the dual attention mechanisms, the \emph{learning to learn} \cite{bertinetto2016learning} module\textemdash
Semantics Induced Learner (SIL)\textemdash is the core component of our framework. SIL is able to learn to fast adapt the predicate classification model conditioned on the semantics dependencies inferred from the categories of subjects and objects, and therefore effectively improves performance of the predicate classification model. The proposed SIL incorporates semantics dependencies into the predicate classification model in a novel and effective way, then dynamically determines, in one-shot manner, the weightings of visual features generated by the bottom-up mechanism, which is challenging to accomplish for a purely static predicate classification model.

\begin{figure}
\includegraphics[width=8cm,height=3cm]{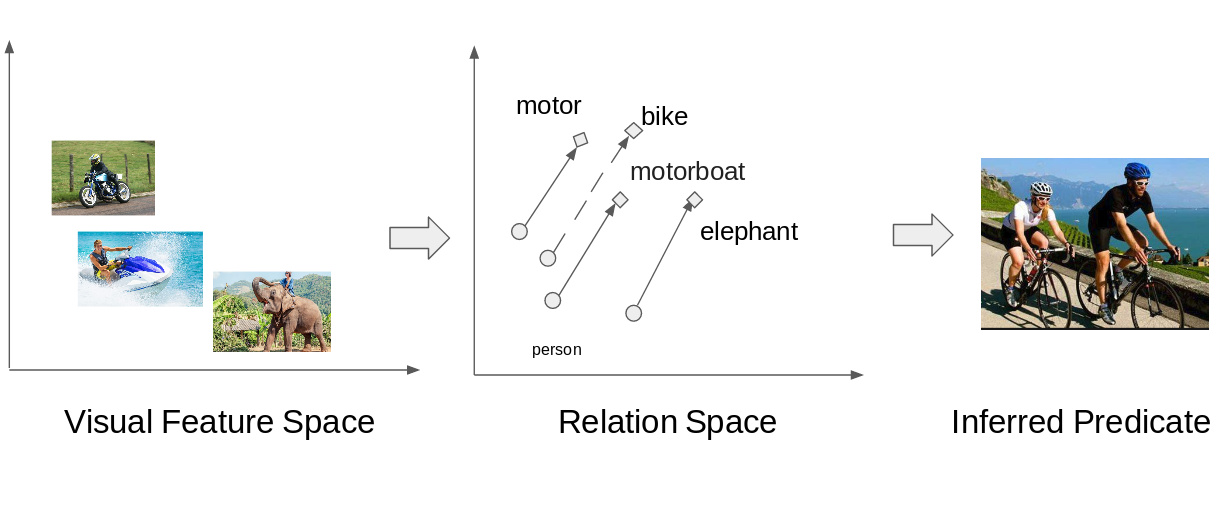}
\caption{An illustration of translation embedding for learning predicate \emph{ride}. Instead of modeling from a variety of ride images, our framework learns consistent translation vector in the relation space regardless of the diverse appearances of subjects (e.g., person) and objects (e.g., horse, motor, etc.) involved in the predicate relation (e.g., ride). With the consistent translation vector, (person, ride, bike) is more likely to be inferred correctly.}
\label{figure3}
\end{figure}

After SIL effectively integrates semantics dependencies (categories of the subject and the object) into the predicate classification model, SIL leverages semantics dependencies  to learn to adapt parameters of the classification model. On the one hand, semantics dependencies can be used to transform the visual features of object relations into a high-dimensional space, where a relation triplet can be split into three parts and mapped as a vector translation. As shown in \ref{figure3}, there are a variety of appearances of the predicate ``ride''. The large variance within the same predicate bring a great difficulty to traditional classification models. However, with the help of the constructed \textbf{semantics translation space}, the only thing that the model needs to learn is the mapping translation between the objects and predicates. This change makes it possible for the classification model to constrain the predicate prediction and relieve the prediction difficulties. On the other hand, with the help of the adaptive parameters generated by SIL, the predicate classification model is able to put more attention on the part of visual features from bottom-up mechanism that have more contributions to predicate classification.

%In particular, SIL can efficiently learn to adapt predicate classification parameters in one-shot manner, achieving fast adaption of the predicate classification model according to specific $semantics$ $dependencies$. We implement SIL by combining a predicate classification network, which we call $Visual$ $Inference$ $Network$(VIN), and a parameter adapter network, which we call $One$-$shot$ $Learner$ $Network$(OLN).

We evaluate the proposed framework on two recently released large datasets for relation detection: Visual Relationship Dataset~\cite{lu2016visual} and Visual Genome~\cite{krishna2017visual}. Experiments show that the SIL module effectively improves the performance for object relation detection to new  state-of-the-art. 

% In summary, our contributions are three-fold: 1) We propose an
% object relation detection module for efficiently learning to adapt
% predicate classification models by exploiting semantics information. 2) We propose a novel method that incorporates semantics
% information into a deep neural network framework more properly.
% 3) Our framework performs better than several strong baselines.
% For example, on two large datasets, the Recall@50 of relation
% predicate recognition are respectively lifted from 47.9\% to 56.56\%
% and from 62.63\% to 66.77\%.

In summary, our contributions are three-fold: 1) We propose an object relation detection moudle (SIL) for efficiently learning to adapt predicate classification models by exploiting semantics information; 2) We propose a novel method that incorporates semantics information into a deep neural network framework more properly; 3) Our framework performs better than several strong baselines. For example, on two large datasets, the Recall@50 of relation predicate recognition are respectively improved from 47.87\% to 56.56\% and from 62.63\% to 66.77\%.

\begin{figure*}
\includegraphics[width=18cm,height=8cm]{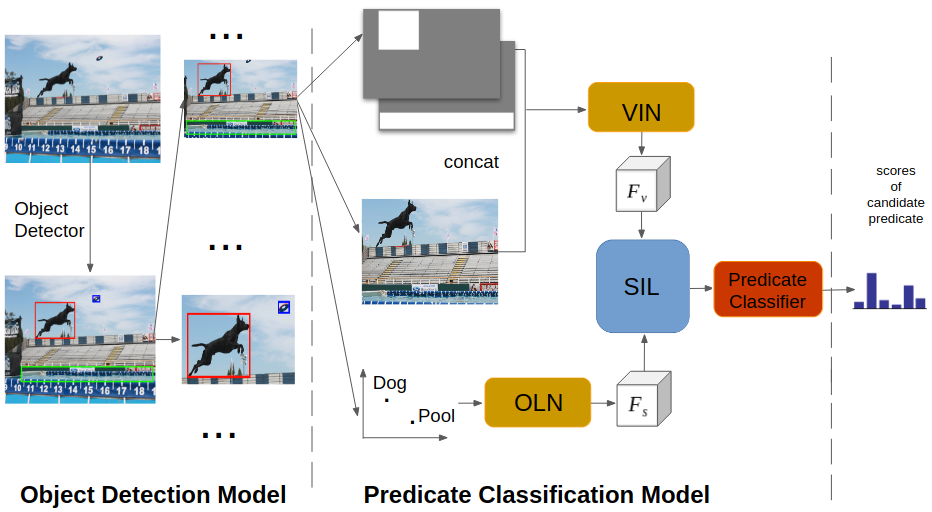}
\caption{Illustration of overall pipeline of the proposed object relation detection framework. Given an image, first an object detector is used to locate individual objects and determine the categories of those objects. The next step is to produce a set of objects pairs from the detected objects. For each pair of objects, the corresponding region cut from the original image and the spatial masks are concatenated, and then fed to \emph{Visual Inference Network} (VIN) to extract visual features $F_v$. Meanwhile, the semantics information(the category vectors of subject and object) is fed to \emph{One-shot Learner Network} (OLN) to generate adaptive parameters $F_s$ in one-shot manner. Taking the $F_v$ and $F_s$ as inputs, the SIL conducts adaptive convolution. Following SIL, a classifier comprising of Fully-connected layers outputs the predicted category probabilities for each kind of predicates.}
\label{mainFigure}
\end{figure*}

\section{RELATED WORK}

With the potential of bridging the semantic gap between images and natural language, object relations in images have been explored in many works in order to make progress in many high-level applications, such as image retrieval~\cite{johnson2015image}, image captioning~\cite{aditya2015images, anderson2016spice} and visual question and answering(VQA)~\cite{wu2017visual}, etc. Different from these attempts which only consider object relation as an intermediate procedure to accomplish those final high-level tasks, we manage to propose an efficient method specifically for object relation detection in this work.

Object relation detection is not a fresh topic. Earliest studies focused on several specific types of relations. \cite{galleguillos2008object} tried to learn four kinds of spatial relation: ``above'', ``below'', ``inside'', and ``around''. \cite{silberman2012indoor} presented a study on the physical support relations between adjacent objects, such as ``behind'' and ``below''. \cite{sadeghi2011recognition, divvala2014learning} regarded object relation detection as a classification task by treating the relation triplet (subject, predicate, object) as a whole. It is worth pointing out that such a strategy would suffer the long-trail problem and Unavoidably fail to detect the variety of infrequent relation types other than a handful of frequent ones with a good number of examples. Besides, all above works utilized handcraft features. Following the conventional idea of splitting the relation triplet into predicate and subject/object two parts, our framework can detect and recognize diverse relation types, such as relative positions (``in front of''), actions (``ride''), functionalities (``part of''), and comparisons (``taller than'').

In recent years, some new advances also have been made for object relation detection. A remarkable family of approaches in which the
relation triplet is regarded as two separable parts, predicate and
subject/object, has become more and more popular. Lu et al. \cite{lu2016visual} formalized object relation detection as a task onto itself and provided a dataset with a moderate number of relation examples. To enable object relation detection on a larger scale, Lu et al. decomposed the relation triplet into two individual parts: predicates and subject/objects.
In \cite{lu2016visual}, sub-images which contain the union of two bounding boxes of object pairs, and language priors, such as the similarity between different relations, are utilized to predict the types of predicates. In \cite{li2017vip, li2017scene}, more attention was put on developing more sophisticated model to extract more representative visual features for object relation detection. \cite{zhang2017visual} predicted predicates simply according to the concatenation of visual appearance, spatial information and category information of objects.

%used the sub-image containing the union of two bounding boxes of object pairs as visual input to predict the predicates and utilized language priors

The approach proposed recently by Dai et al.\cite{dai2017detecting} is the most related work to our framework. In \cite{dai2017detecting}, Dai et al. proposed a particular form of RNN based on CRF to exploit the statistical dependencies among predicates, subjects, and objects, while refining the estimates of posterior probabilities iteratively. Our method differs from them in two aspects. First, our method incorporate the semantic information into a deep neural network in a more simple and effective fashion instead of iteratively fusing them into the predicate classification model. Second, our framework exploits the semantics information, dynamically determines visual feature weightings and enables a combination of a bottom-up attention mechanism and a top-down mechanism for object relation detection.

\section{METHOD}
In this section, we elaborate on the proposed method for object relation detection. We start with an overview of the proposed pipeline in Sec. 3.1, followed by a description of three modalities utilized for relation prediction in Sec. 3.2. Then, Sec. 3.3, Sec. 3.4 and Sec. 3.5 give a detailed introduction to the object detection module, relation classification module and SIL Module, respectively.

%\subsection{Type Changes and {\itshape Special} Characters}
\subsection{Pipeline Overview}
There are two different strategies for object relation detection: one is to consider each relation triplet as a distinct relation type, the other is to recognize each component of the relation triplet respectively. As discussed above, the former is not practically feasible for most tasks, due to the challenge from the large number of relation types and the desperately imbalanced distribution within them. In this work, we follow the second strategy and manage to lift the performance to a new level. To be more specific, we dedicate to developing a new framework that can effectively capture the rich information in images (including not only appearance information, but also spatial and semantic information) and take advantage of the information in an associated way to facilitate the object relation detection. Our framework is comprised of two parts: object detection model and predicate classification model. The pipeline of our method is illustrated in \ref{mainFigure}.

%is built on 

\textbf{(1) Object detection Model}.  Our framework starts from an object detection model composed of a region proposal network (RPN) and a classification network. In particular, we take advantage of Faster-RCNN due to its state-of-the-art performance. Given an image, Faster R-CNN is utilized to detect a set of candidate object proposals. Each candidate object proposal comes with a bounding box and a classification confidence score.

\textbf{(2) Predicate classification Model}. The next step is to form a set of object pairs from the detected object proposals. With $m$ detected object proposals output from Faster R-CNN, $m(m - 1)$ pairs will be generated. A triplet, (subject, predicate, object), will be output after these pairs of object proposals, including their appearance information, spatial information and category information, are fed to the relation classification module. The predicate classification model comprises of 2 components: a \emph{Visual Inference Network} for extracting visual and spatial features; a \emph{One-shot Learner Network} for exploiting semantics information.

\subsection{Modalities for Predicate Classification}
In predicate classification module, three modalities are taken into consideration, which are listed in detail below.

\textbf{(1) Semantics dependencies}. In a triplet (s, p, o), there exist strong semantics dependencies between the relation predicate $r$ and the object categories s and o. For example, (person, play, piano) is common, while (piano, play, person) is totally unlikely. Given any valid relation, (s, p, o) can be represented in a high-dimensional vector $s, p$, and $o$, respectively, thus the relation is represented as a translation in the high-dimensional space: $s + p \approx o$ when the relation holds. Semantics dependencies can facilitate the predicate classification model in learning consistent mapping translation in the high-dimensional space regardless of the diverse appearances of subjects and objects. So exploiting semantics information is a simple yet effective way to alleviate the $intra$-$class$ $divergence$ problem. See the \ref{figure3}. 

At the same time, we implement the top-down attention mechanism for predicate classification by inputting semantics dependencies to SIL, which will be detailed in the next subsection. The top-down attention mechanism takes advantage of object-specific information to predict an attention distribution over the visual features from the bottom-up mechanism.

\textbf{(2) Appearance}. As mentioned above, each detected object proposal comes with an bounding box to indicate its spatial information, which can be used to extract its visual features and thus infer its category( sometimes the category output by Faster R-CNN is not correct). In addition, the type of the relation may also be reasoned visually. In order to exploit this information, we utilize a CNN \cite{simonyan2014very} to sub-images, which contain the union of two bounding boxes of object proposal pairs with a small margin, to extract appearance features. In this case, the appearance extracted from the enclosing box exploits not only the representations of object proposals themselves but also the ones of the surrounding context. 

%a bounding box that encompasses both objects with a small margin

\textbf{(3) Spatial Information}. The spatial configurations between two object proposals may be also helpful for inferring the predicate type between them, such as the relative positions and sizes of two object proposals. Such information can be complementary especially when the appearance information is unqualified, e.g. due to photometric variations.

But there comes a question: how should we represent and input
this spatial information to our predicate classification model to leverage it
in a better way? The practice of geometric measurement \cite{johnson2015image} is simple, however it may unavoidably ignore certain information of the configurations. In this work,
we utilize dual spatial masks as the representation, which consist of  two binary masks, one for the subject and the other for the object.
We generate the masks, which might overlap with each other, from
the bounding boxes of the object proposals, as shown in \ref{mainFigure}.
The difference between our work and previous works is that we concatenate the
two-channel masks with the sub-image to generate a five-channel
``image''. Concatenating the masks and the sub-image helps the CNN
learn the relation between these two different modalities of inputs
in a more unified and implicit way. In this case, we implement
bottom-up attention using Faster R-CNN, which represents a natural expression of a bottom-up attention mechanism.

\begin{figure}
\includegraphics[width=8cm,height=4cm]{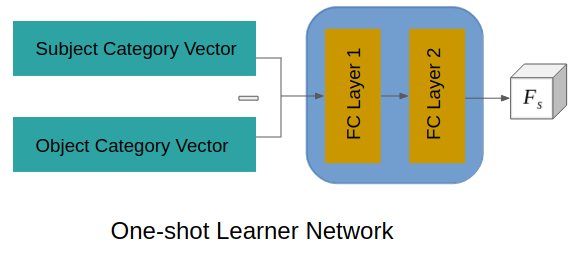}
\caption{Illustration of the architecture of the One-shot Learner Network. The One-shot Learner Network takes as input the category information (the ``Subject Category Vector" element-wisely subtracts the ``Object Category Vector") and outputs the adaptive convolution parameters $F_s$. The One-shot Learner Network is composed of 2 fully-connected layers. In particular, the output size of the fully-connected layer 1 is 300, and the output size of fully-connected layer 2 is 256.}
\label{figure5}
\end{figure}

\subsection{Object Detection Module}

For object detection module, we take advantage of the Faster-RCNN object detection network with the VGG-16 architecture.  More implementation details will be illustrated in Experiment Section.

\subsection{Predicate Classification Module}

Predicate classification module is comprised of three parts: the visual inference network, the one-shot learner network and a simple predicate classifier.

\textbf{Visual Inference Network}.  Visual inference network is mainly based on VGG-16 architecture (deleting fully-connected layers) due to its high performance in feature extracting and exploiting. Visual inference network takes in the concatenation of spatial information and semantics dependencies. See Figure \ref{mainFigure}. Such concatenation can enable bottom-up attention in a more instinctive and unified way, thus visual inference network can capture rich relation representations.

\textbf{One-Shot Learner Network}. It is essentially a parameter adapter which takes in the semantics dependencies to yield the dynamic parameters $F_s$ for the Visual Inference Network. We implement it by a small network consisting of two fully-connected layers. Its architecture is shown in Figure \ref{figure5}. In particular, the tensor $F_s \in R ^ {h \times w \times c}$ generated from the last layer of One-shot Learner Network is taken as the dynamic convolutional kernels of the Visual Inference Network. Here convolution kernel size $h$ and $w$ is $1$, and the number of channels to learn for dynamic convolution is $c$, where $c$ is the number of output channels of last fully-connected layer of One-shot Learner Network. 

\textbf{Predicate Classifier.} The predicate classifier consists of three fully-connected layers. And drop-out layers are used to avoid over-fitting.

\subsection{Semantics Induced Learner}

\begin{figure}
\includegraphics[width=8cm,height=6cm]{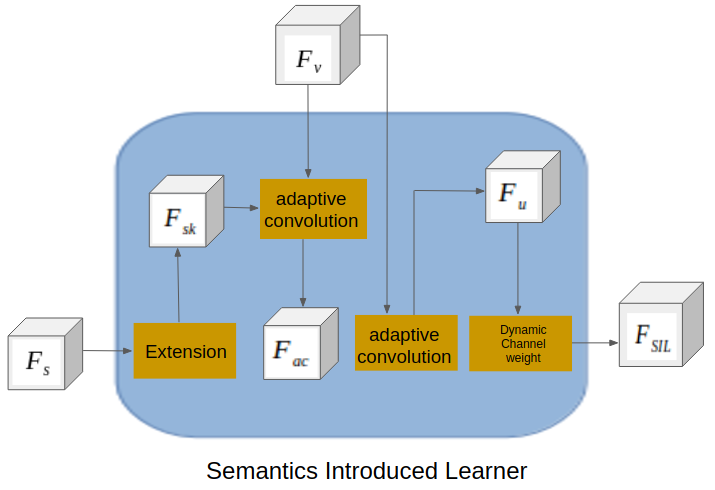}
\caption{The illustration of the Semantics Induced Learner. The SIL takes the semantics feature from Semantics Network, $F_s$, and the visual feature from Visual Network, $F_v$, as input. Then $F_s$ is extended to the same channel with $F_v$. After conducting adaptive convolution with $F_s$ and $F_v$, the $F_u$ is generated by concatenating the output of adaptive convolution with $F_v$. Finally, dynamic weights are given to different channels of $F_u$, and SIL produces $F_{SIL}$.}
\label{figure6}
\end{figure}

As shown in Figure \ref{figure6}, we use Semantics Induced Learner module to connect visual inference network, one-shot learner network and the predicate classifier. In this section, we will illustrate the SIL module in details. Basically, SIL module is comprised of three operations: \textbf{Extension}, \textbf{Dynamic Convolution}, \textbf{concatenation}. The details are list below.

\textbf{Extension}. Traditionally, $c$, the number of channels of $F_{s}$, should be $c_i \times c_o$, where $c_i$ and $c_o$ is the number of input and output channels of dynamic convolution, respectively. However, it is ill-suited for one-shot learner network to totally yield all convolution kernels because the cost resulting from those excessively large quantity of parameters ( kernels) is not affordable. For example, for such a dynamic convolution where input feature maps are 512 channels and output feature maps are 256 channels, the number of convolution kernels to be predicted by the One-shot Learner Network is as large as $512 \times 256$. This would cause unbearable high space and time cost, at the same time, the information carried by $semantics$ $dependencies$ is not enough to be transformed into such a huge amount of parameters. To avoid these issues, we yield dynamic convolutional kernels by extending $F_s$ to $c_i$ channels.

\begin{equation}
  F_{sk} = E(F_s, n)
\end{equation}where $F_{sk}$ are dynamic convolutional kernels, $E$ denotes the extension operation and $n$ is the number of extension.

\textbf{Dynamic Convolution}. To take full advantage of the extended dynamic convolutional kernels $F_{sk}$ for extracting the most useful features for the predicate classification model, we apply $F_{sk}$ on the high-level features $F_v$ output from the Visual Inference Network. In other words, we implement the top-down attention mechanism via a dynamic convolution layer. There is no remarkable difference between the dynamic convolution layer and traditional convolution layers, except that the static convolution kernels are replaced by the predicted dynamic convolution kernels $F_{sk}$.

It is worth mentioning out that the difference between features extracted by traditional static CNN and those features extracted a dynamic convolution layer is that the latter are generated by the dynamic kernels $F_{sk}$, which are injected with semantics information of a given pair of objects, instead of purely hand-crafted convolutional kernels. In this way, we subtly incorporate the semantics information into a deep neural network. Moreover, the classification model will be forced to specifically pay attention to useful feature regions according to different object pairs, because the kernels of the dynamic convolution layer $F_{sk}$ are efficiently learned by the One-shot Learner Network with semantics dependencies as input. In this case, we enable the predicate classification module to learn diverse object relations according to different pairs of objects in one-shot manner.

\begin{equation}
  F_{ac} = K(F_{sk}, F_v)
\end{equation}where $K$ denotes a traditional convolution operation and $F_{ac}$ are features extracted the dynamic convolution layer.

%adapting diverse appearances of objects involved in the predicate relation.

% Different from the features $F_v$ extracted from the Visual Inference Network, $F_{ac}$ is integrated with
% semantics information and thus complementary to $F_v$ for predicate
% classification. Consequently, after outputting $F_u$ by concatenating
% $F_v$ and $F_{ac}$, we obtain $F_{SIL}$, the final features for predicate classification, by adopting the dynamic channel weight   to
% adaptively recalibrate channel-wise feature responses.

\textbf{Concatenation}. Different from features $F_v$ extracted from the Visual Inference Network, $F_{ac}$ is integrated with semantics information and thus complementary to $F_v$ for predicate classification. Consequently, after outputting concatenating $F_v$ and $F_{ac}$, we obtain the final features for predicate classification, by adopting the dynamic channel weight idea \cite{hu2017squeeze} to adaptively recalibrate channel-wise feature responses.

\begin{equation}
\begin{split}
  &F_{u}  = C(F_{ac}, F_v) \\
  &F_{SIL} = D(F_u)
\end{split}
\end{equation} where $C$ denotes a concatenation operation and $F_{SIL}$ denotes a dynamic channel weight operation. $F_u$ is the concatenating result from $F_{ac}$ and $F_v$, and $F_{SIL}$ is the final features for the predicate classifier.

%adapting diverse appearances of objects involved in the predicate relation.

%by explicitly modelling interdependencies between channels.

\section{Experiments}

\textbf{Dataset}. We evaluate our model on two large scale datasets for object relation detection: \textbf{(1) Visual Relationship Dataset (VRD).} This dataset is proposed in \cite{lu2016visual}, which consists of 5,000 images with 100 object categories and 70 predicates, with 4,000 images for training  and the remaining for test. In total, the dataset contains 37,993 relation examples that belong to 6,672 relation types and 24.25 predicates per object category on average. In the experiments, we adopt the same split with \cite{lu2016visual}. \textbf{(2) Visual Genome (VG).}: 
Annotations of the original Visual Genome \cite{krishna2017visual} dataset are noisy, so some data pruning over the official revision are necessary. For instance, both ``young woman'' and ``lady'' are hyponymy to the ``woman'' in natural language. Therefore, in \cite{zhang2017visual}, Zhang et al. constructed a subset of the original Visual Genome dataset, which consists of 99,651 images with 200 object categories and 100 predicates, by filtering out relations with less than 5
samples. In particular, the subset contains 1,174,692 relations examples that belong to 6,672 relation types and 57 predicates per object category on average. We follow the split in \cite{zhang2017visual}, namely using 73,793 images for training and 25,858 images for test.

\begin{table*}
  \caption{Comparison with baseline models, with $Recall@50$ and $Recall100$ as the metrics. ``-'' is used to denote not applicable. For instance, VP regards a relation triplet as a whole to detect, giving no results on Two Boxes Detection task. Note that all methods above only output one predicate with the highest confidence score for one object pairs.}
  \label{table1}
  \begin{tabular}{cccccccccccccccccccccccccccccccccccccccccccccccccccc}
    \toprule
   \multirow{2}{*}{} & \multicolumn{1}{c}{} &\multicolumn{2}{c}{Predicate Classification} & \multicolumn{2}{c}{Union Box Detection} & \multicolumn{2}{c}{Two Boxes Detection} \\
    \cmidrule[\heavyrulewidth](lr){3-4}  \cmidrule[\heavyrulewidth](lr){5-6}\cmidrule[\heavyrulewidth](lr){7-8} 
  & & Recall@50 & Recall@100  & Recall@50 & Recall@100 & Recall@50 & Recall@100 \\
    \midrule    
   \multirow{5}{*}{VRD}  & VP~\cite{sadeghi2011recognition} & 0.97 & 1.91 & 0.54 & 0.63 & - & - \\  
    & Joint-CNN~\cite{fang2015captions} & 1.47 & 2.03 & 0.07 & 0.09 & 0.07 & 0.09 \\
    & VR~\cite{lu2016visual} & 47.87 & 47.87 & 16.17 & 17.03 & 13.86 & 14.70 \\
    & VTranE~\cite{zhang2017visual} & 44.76 & 44.76 & 19.42 & 22.42 & \textbf{14.07} & 15.20 \\
    & Ours & \textbf{56.56} & \textbf{56.56} & \textbf{20.82} & \textbf{24.50} & 13.81 & \textbf{16.01} \\
    
    \midrule       
   \multirow{5}{*}{VG}  & VP~\cite{sadeghi2011recognition} & 0.63 & 0.87 & 3.41 & 4.27 & - & - \\  
    & Joint-CNN~\cite{fang2015captions} & 3.06 & 3.99 & 1.24 & 1.60 & 1.21 & 1.58 \\
    & VR~\cite{lu2016visual} & - & - & - & - & - & - \\
    & VTranE~\cite{zhang2017visual} & 62.63 & 62.87 & 9.46 & 10.45 & 5.52 & 6.04 \\
    & Ours & \textbf{68.63} & \textbf{68.91} & \textbf{10.60} & \textbf{12.05} & \textbf{5.96} & \textbf{6.64} \\
    
   \bottomrule
  \end{tabular}
\end{table*}

\textbf{Implementation details} For the object detection module, at training time, the mini-batch contains 256 region proposal boxes generated by the RPN of Faster-RCNN. Those proposal boxes are positive if it has an intersection over union (IoU) of at least 0.7 with some ground truth regions and it is negative if the IoU < 0.3. The positive proposals are fed into the classification network (RCNN), where each proposal outputs an $(N + 1)$ class confidence score and $N$ bounding box estimations. Next, non-maximum suppression (NMS) is performed for every class with the IoU > 0.4, resulting in 15.6 detected objects on average, each of which has only one bounding box. At test time, 300 proposal regions generated by RPN with IoU > 0.7 are sampled. Following the classification network (RCNN), NMS is still performed with IoU > 0.6 on the those 300 proposals output by RCNN, resulting in 15–20 detections per image on average. The Faster-RCNN is initialized with model pre-trained on ImageNet. 

For the predicate classification module,  when training for \emph{Union Box Detection} and \emph{Two Boxes Detection} tasks, any object generated by Faster RCNN is regard as positive if it has an intersection over union (IoU) of at least 0.7 with some ground truth objects and it is negative otherwise. At test time, all pairs produced by objects with its confidence score of at least 0.1 from Faster RCNN are fed into predicate classification module to classify the predicate. We randomly initialize the relation detection network with Xaiver weights.

\subsection{Experiment Settings}

\textbf{Performance metrics.} Following \cite{lu2016visual}, we use Recall@50 (R@50) and Recall@100 (R@100) as the performance metrics in our experiments. Recall@K is the fraction of ground-truth instances that are correctly recalled in top K predictions in an image. It is worth pointing out that the annotations in the datasets above are incomplete. So using average precision as metrics will penalize the detection once there is not the particular ground truth.

\textbf{Task settings.} We split a relation triplet into two components: objects and predicate. So in order to detect object relations, we need localize and classify objects, then predict predicate. Following~\cite{lu2016visual}, to study the performance of our model on these tasks, we evaluate our framework in the following settings:

\textbf{(1) Predicate classification.} This task is to purely predict predicates between object pairs. It allows us to study the performance of our model without the limitations of object detection. The categories and locations of both the subject and the object are given. 

\textbf{(2) Union box detection.} In this task, nothing but an image is given. The entire relation is treated as a bounding box. A detection is considered correct if all three
parts of the triplet (s, r, o) are correctly classified, and simultaneously the predicted bounding box has at least 0.5 overlap with ground truth box. 

\textbf{(3) Two boxes detection.} This task  also only one image is provided. A prediction is considered correct if all three parts of the triplet (s, r, o) are correctly classified, and simultaneously the predicted bounding boxes of subject and object have at least 0.5 overlap with ground truth boxes, respectively. This task is more difficult than \emph{Union Box Detection}  task.

\subsection{Comparative Results}

\textbf{Baseline models}. We compared our model with state-of-the-art ones under the above three task settings
above. \textbf{(1) Visual Phrase (VP)}~\cite{sadeghi2011recognition}: a typical method that regards a triplet (subject, predicate, object) as a whole to classify. The object detector is DPM object detection model~\cite{felzenszwalb2010object}. \textbf{(2) Joint-CNN}~\cite{fang2015captions}: a joint approach where the types of subject, object and predicate are simultaneously classified by a deep neural network. \textbf{(3) Visual Relationship (VR)}~\cite{lu2016visual}: This work uses R-CNN to detect objects, then constructs separate visual models for classifying objects and predicates, later refines the likelihood of predicates by leveraging language priors from semantic
word embeddings. \textbf{(4) VTranE}~\cite{zhang2017visual}: an end-to-end relation detection network that enables object-relation knowledge transfer in a fully-convolutional fashion. Training and testing are in a single forward/backward pass.
%\ref{table1}

Comparison of results with baseline models on VRD and Visual Genome datasets are listed in Table 1. From Table 1, we can find that: (1) The performances of VP~\cite{sadeghi2011recognition} and Joint-CNN~\cite{fang2015captions} is so poor on the two datasets. VP~\cite{sadeghi2011recognition} regards a relation triplet as a whole to classify. Its poor performance indicates that such strategy cannot enable models to learn such a great number of relation types with imbalanced examples. (2) The results of Joint-CNN~\cite{fang2015captions} indicates that it is impractical for a CNN to learn feature representations for both predicates and objects simultaneously. (3) VR~\cite{lu2016visual}
and \cite{zhang2017visual} obtain remarkable performance improvements compared with VP~\cite{sadeghi2011recognition} and Joint-CNN~\cite{fang2015captions}. However, their predicate classification models are not effective yet. In particular, their approach to extract and exploit visual and semantics features is not robust. So their performances are not ideal. (4) Our
method outperforms the state-of-the-art framework~\cite{zhang2017visual} by a
remarkable maigin in the predicate classification task, e.g. from 44.76 to 56.56 on VRD and from 62.63 to 68.63 on VG (with Recall@50 as metrics). Such remarkable performance gains indicates the effectiveness of our predicate classification model. (5) However, the progress on Union box detection and two boxes detection tasks is not satisfactory yet. Due to the incomplete annotations of the two datasets, it is difficult to train an effective and robust object detector.

\subsection{Ablation study}

Experiments on the two large scale datasets demonstrate the effectiveness of our predicate classification network. Figure 7 lists some qualitative results of our method. To confirm the performance improvements of our framework are due to the Semantics Induced Learner module, we conduct necessary ablation study on VRD and VG. We construct a controlled framework (CF) by deleting the SIL module from our original framework. And the comparison of results of such two frameworks on three tasks is listed in Table 2.

From Table 2, we can see the performances of framework with SIL module is far better than the ones of the controlled framework on both datasets. So we can confirm the effectiveness of our proposed Semantics Induced Learner module for object relation detection.

\begin{table*}
  \caption{Ablation study of our model.}
  \label{table2}
  \begin{tabular}{cccccccccccccccccccccccccccccccccccccccccccccccccccc}
    \toprule
   \multirow{2}{*}{} & \multicolumn{1}{c}{} &\multicolumn{2}{c}{Predicate Classification} & \multicolumn{2}{c}{Union Box Detection} & \multicolumn{2}{c}{Two Boxes Detection} \\
    \cmidrule[\heavyrulewidth](lr){3-4}  \cmidrule[\heavyrulewidth](lr){5-6}\cmidrule[\heavyrulewidth](lr){7-8} 
  & & Recall@50 & Recall@100  & Recall@50 & Recall@100 & Recall@50 & Recall@100 \\
    \midrule    
   \multirow{2}{*}{VRD}  & CF & 46.60 & 46.60 & 6.51 & 7.96 & 4.27 & 5.3 \\  
    & Ours & \textbf{56.56} & \textbf{56.56} & \textbf{20.82} & \textbf{24.50} & \textbf{13.81} & \textbf{16.01} \\
    \midrule    
   \multirow{2}{*}{VG}  & CF &52.13 & 52.32 & 2.61 & 3.13 & 0.98 & 1.43 \\  
    & Ours & \textbf{68.63} & \textbf{68.63} & \textbf{10.60} & \textbf{12.05} & \textbf{5.96} & \textbf{6.64} \\
    
   \bottomrule
  \end{tabular}
\end{table*}

\subsection{Discussion}

From the quantitative results in Table 1, we have four observations:
\begin{figure}
\includegraphics[width=8cm,height=10cm]{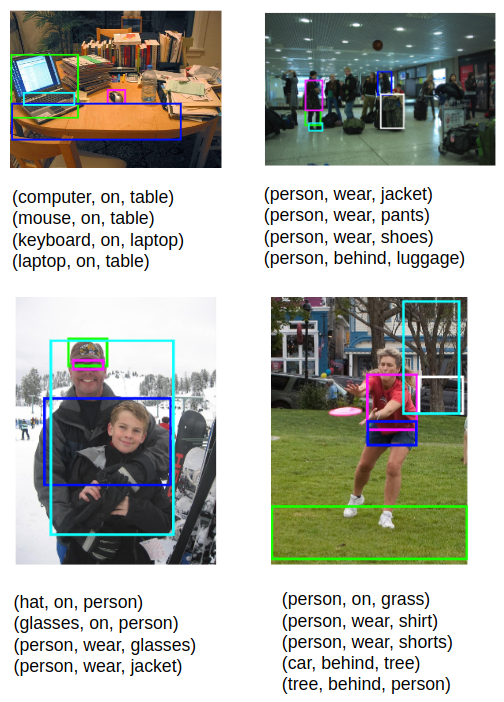}
\caption{Some qualitative results of our framework.}

\end{figure}
(1) The strategy that treats an object relation triplet as two separate parts is more suitable for the object relation detection task.  The frameworks which develop two models for object detection task and predicate classification task respectively, such as VR and VTranE, obtain remarkable performance improvements over the ones that only have an unified model for the whole object relation detection task, e.g. VP. Because if a relation triplet is considered as a whole, the number of types of relations would become excessively large. Moreover, due to imbalanced distribution within datasets, there is no enough examples for learning a majority of rare types of relations.

\begin{table}[!h]
  \caption{The mean Average Precisions of VRD and VG are listed below. Such poor results are mainly due to the incomplete annotations of datasets. When we utilize two separate models to detect object relations, an object detector with relatively good performances is necessary for obtaining satisfactory relation detection results.}
  \label{table3}
  \begin{tabular}{cccccccccccccccccccccccccccccccccccccccccccccccccccc}
    \toprule
 &    VRD & VG \\
      \midrule    
 mAP &   18.37 & 9.62\\
   \bottomrule
  \end{tabular}
\end{table}

(2) The performance of object detectors has a great influence upon the performance of the object relation detection task. Even if our framework achieves a relatively high recall on predicate classification, which indicates the predicate classification model is qualified, the performances on the other two tasks are still poor due to the unsatisfactory performances of object detectors. We list the performances of our Faster RCNN on both datasets in Table 3. Basically, the incomplete annotations in two datasets are the main reason why Faster RCNN can not achieve a good mAP.

(3) The way in which different modalities are exploited and incorporated plays an important role in the predicate classification task. In generally, VTranE  and our framework are fed into the same modalities: appearance, spatial information and objects categories. Firstly, even if we only utilize appearance and spatial information, we obtain better performance than the one of VTranE on VRD. This indicates that concatenation of spatial masks with a image is more beneficial for models to capture implicit representations of relations. On the other hand, our framework captures the characteristic of predicate classification task - the categories of objects play important roles in a relation triplet. The way in which semantics information of objects is utilized in SIL enables not only a top-down attention mechanism, but also the models to learn in one-shot manner. Both factors facilitate the performance improvements of our framework in predicate classification task.

\section{Conclusion}
	
We pay attention to the object relation detection task in this work, which has great potential in image understanding. We propose a new framework for object relation detection, which consists of an object detection model and a predicate classification model. The predicate classification model integrates three modalities: semantics dependencies, appearance and spatial information. The core of the predicate classification model is the Semantics Induced Learner module, which subtly  incorporates semantics dependencies into the predicate classification model and enables the predicate classification model to predict object relations in one-shot manner. Our experiments on both Visual Relationship Dataset and Visual Genome dataset show the effectiveness and robustness of SIL module for detecting diverse object relations.

\vspace{-2mm}
\section*{Acknowledgement}
\vspace{-1mm}

The work of Jiashi Feng was partially supported by NUS startup R-263-000-C08-133, MOE Tier-I R-263-000-C21-112, NUS IDS R-263-000-C67-646 and ECRA R-263-000-C87-133.

{\small
\bibliographystyle{ieee}
\bibliography{my_citation}
}

\end{document}